\documentclass[10pt,conference]{IEEEtran}
\IEEEoverridecommandlockouts
\usepackage{amsmath,graphicx,amssymb,subcaption,booktabs, multirow,amsthm}

\usepackage{setspace}

\title{Peer-to-Peer Learning+Consensus\,with\,Non-IID Data}
\author{\IEEEauthorblockN{Srinivasa Pranav}
\IEEEauthorblockA{\textit{Electrical and Computer Engineering} \\ \textit{Carnegie Mellon University } \\ Pittsburgh, USA \\
spranav@cmu.edu}
\and
\IEEEauthorblockN{Jos\'e M.F. Moura}
\IEEEauthorblockA{\textit{Electrical and Computer Engineering} \\ \textit{Carnegie Mellon University } \\ Pittsburgh, USA \\
moura@andrew.cmu.edu}
\thanks{
This work was partially supported by NSF Grant CCF-2327905 and XSEDE \cite{6866038}
Allocation ELE220003
on PSC Bridges-2 \cite{10.1145/3437359.3465593}. First author partially supported by NSF Graduate Research Fellowship (GRFP; Grants DGE1745016, DGE2140739), and an ARCS Fellowship. Authors thank Tyler Vuong and Shreyas Chaudhari for insightful discussions.
}
}
\begin{document}
%
\maketitle
%
\begin{abstract}
Peer-to-peer deep learning algorithms are enabling distributed edge devices to collaboratively train deep neural networks without exchanging raw training data or relying on a central server. Peer-to-Peer Learning (P2PL) and other algorithms based on Distributed Local-Update Stochastic/mini-batch Gradient Descent (local DSGD) rely on interleaving epochs of training with distributed consensus steps. This process leads to model parameter drift/divergence amongst participating devices in both IID and non-IID settings. We observe that model drift results in significant oscillations in test performance evaluated after local training and consensus phases. We then identify factors that amplify performance oscillations and demonstrate that our novel approach, P2PL with Affinity, dampens test performance oscillations in non-IID settings without incurring any additional communication cost.
\end{abstract}


\section{Introduction}
The proliferation of the Internet-of-Things has led to a deluge of smart devices that collect, store, and process data at the network edge. These smart devices each collect relatively small amounts of data and are interested in performing inference tasks like image classification. Recent hardware advancements \cite{5Gsurvey, Camaroptera, deeplearningonmobiledevices} are enabling smart devices to achieve high test performance by training neural networks on-device. These trends make peer-to-peer deep learning approaches a crucial component of evolving 6G environments.

We consider the peer-to-peer setting where distributed devices wirelessly collaborate with nearby devices (neighbors) to train neural networks and accomplish a single task \cite{pranav2023peer, Italians, wang2021cooperative}. Since devices' local datasets are relatively small in practice, attaining high test performance relies on collaboration among devices. Unlike the federated learning setting \cite{mcmahan2017communication}, there is no central server or device to facilitate collaboration among devices. The peer-to-peer learning setting instead leverages device-to-device communication to catalyze the training process and models the network as a communication graph. Motivated by privacy concerns and the European Union's GDPR legislation, we examine settings where devices exchange trained deep model parameters instead of raw training data.

The peer-to-peer learning problem can be formulated as a distributed finite-sum optimization problem that is nonconvex and requires global consensus in the limit. When data at individual devices are homogeneous or independently and identically drawn from the same distribution (IID), the local training loss (objective) functions are similar to each other. However, heterogeneous (non-IID) data at each device leads to disparate local objectives and new optimization challenges.

\begin{figure}[h]
    \centering
    \includegraphics[width=0.25\columnwidth]{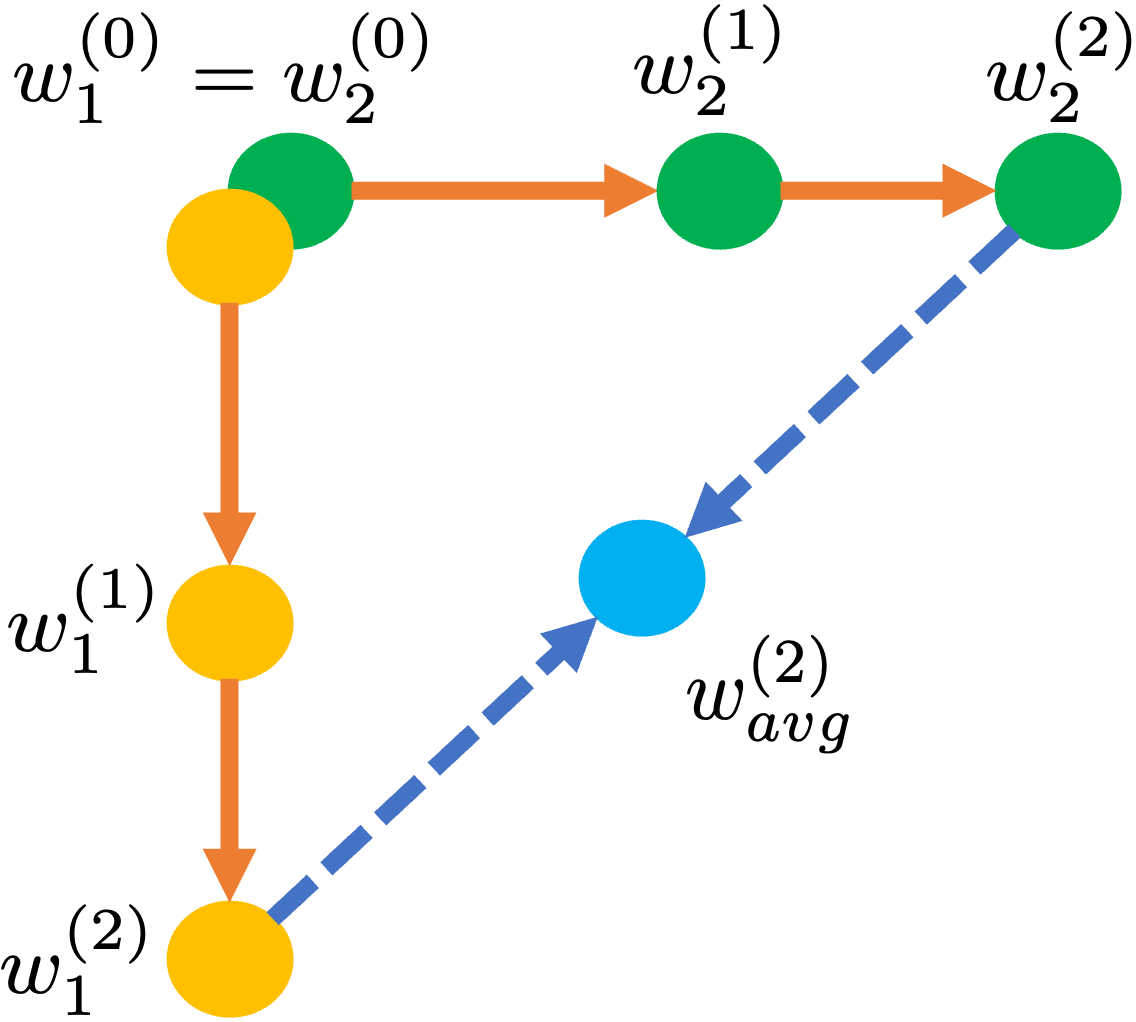}
    \caption{\fontsize{9}{10.5}\selectfont Example of model drift/divergence caused by non-IID data.}
    \label{fig:model_drift}
\end{figure}

The most popular algorithms for the peer-to-peer setting avoid sharing local data and are variants of distributed periodic/local-update stochastic/mini-batch gradient descent (local DSGD) \cite{pranav2023peer, Italians, koloskova2020unified, jiang2017collaborative, lian2017can}. Peer-to-peer learning algorithms interleave epochs of local learning with average consensus steps that combine gradient information between nearby devices. During local learning, devices independently extract knowledge from their local datasets by performing multiple gradient descent updates according to their local loss functions. This process causes different devices' model parameters to ``drift" apart. During the average consensus phase, devices share newly obtained knowledge by averaging their updated model parameters with their neighbors' parameters. This process brings parameters closer together (see Fig.~\ref{fig:model_drift}).

\textbf{Contributions:}
In this paper, we show that drift or divergence in model parameters during local training (and the correction from average consensus) surprisingly causes significant oscillations in test performance evaluated after the local training phase and after the consensus phase.

We highlight that test accuracy oscillations are sizeable even when devices have homogeneous (IID) data and are connected by a complete graph (8\% at start and 0.5\% near convergence). Despite these oscillations in test performance, P2PL's \cite{pranav2023peer} performance evaluated after consensus steadily increases. 

The sawtooth-like oscillations are amplified by heterogeneous (non-IID) data. These larger oscillations are present even when only two participating devices learn to classify data from classes they never saw during training. Local training drastically decreases performance on unseen classes while improving performance on seen classes. Meanwhile, consensus decreases performance on seen data while drastically increasing performance on unseen data.

We further empirically characterize how performance oscillations from model drift and consensus are amplified by the number of local gradient updates and complexity of the task.

Finally, we propose P2PL with Affinity, an algorithm based on P2PL, and demonstrate that it dampens test performance oscillations in non-IID settings without incurring any additional communication cost.




\section{Related Work}
Gradient tracking methods \cite{soummyaGT} were developed for distributed convex optimization problems where local objective functions do not share the same minimum as the global objective. However, they requires expensive communication after every gradient step and generally only have asymptotic guarantees reaching the minimum of the global objective. In contrast, we focus on communication-efficient nonconvex distributed optimization with multiple local gradient steps and non-IID data that cause large oscillations in test performance early in the training process.

Model/client drift was also studied in federated learning settings \cite{zhao2018federatednoniid, karimireddy2020scaffold}, where heterogenous (non-IID) data amplified model drift. These works focused on parameters divergence, but we observe and focus on the effect of model drift: oscillations in test performance. Furthermore, the server in federated learning enforces global consensus in every round while we study how peer-to-peer distributed consensus affects drift and osscillations in test performance over time (since model drift disappears only in the limit). 

Theoretical works \cite{jiang2017collaborative, koloskova2020unified, karimireddy2020scaffold} often bound model drift by assuming bounded gradient dissimilarity and relying on smoothness, but both of these assumptions may not hold in practice. Our experiments consider pathological non-IID data that increases gradient dissimilarity since devices only see and train with a subset of classes.

In Ref.~\cite{pranav2023peer}, we introduced the Peer-to-Peer Learning (P2PL) algorithm, which empirically improved on local DSGD to be more communication efficient. In this paper, we empirically show that it achieves good test performance despite the fact that it also experiences oscillations in test performance.


\section{Problem Formulation}
\subsection{Optimization Problem}
We are interested in minimizing the following weighted finite-sum optimization objective, where $\sum_k n_k = n$:
\begin{equation}
    \min_{w} \frac{1}{n} \sum_{k=1}^K n_k F_k(w)
    \label{eq:objective}
\end{equation}
In the peer-to-peer setting, the $K$ terms of the summation correspond to $K$ different distributed devices. The $k$th device has its own corresponding state or parameter vector, $w_k$, and can only access its own local objective function $F_k$. 

We focus on peer-to-peer deep learning, where $w_k$ is the $k$th device's vectorized deep neural network parameters and $F_k$ is the $k$th device's \textbf{nonconvex}, differentiable local (empirical) training loss function. With $n_k$ being the size of the $k$th local training dataset, the distributed objective is equivalent to the empirical training loss obtained in the case of centralized data and training (the cloud/datacenter setting).

The goal of training is to achieve good generalization performance, which we define as achieving high test performance on a test dataset that contains samples drawn from the same distribution as the training data. Communication among devices is necessary for generalization since, in practice, the data collected at an individual device is often insufficient for training a deep model to generalize well. 

\subsection{Consensus and IID vs. Non-IID Data Distributions}
We also require the distributed devices' model parameters to converge to the same local minimum of the global objective function in the limit. Letting $w_k^{(t)}$ be the $k$th device's model parameters at round $t$ of an iterative optimization algorithm, we define global consensus as:
\begin{equation}
    \forall k\in[K],\;\; \lim_{t\to\infty} w_k^{(t)} = w^{(\infty)}
    \label{eq:consensus}
\end{equation}
Achieving \eqref{eq:consensus} requires inter-device communication.

When data at the devices are homogenous or independently and identically drawn from the same distribution (IID), the nonconvex local training loss (objective) functions are similar to each other and, in expectation, have the same local minima. While this makes global consensus \eqref{eq:consensus} easier to achieve, the use of empirical loss functions and stochastic function (and gradient) evaluation can still slow convergence. 

When data at the devices are heterogeneous and not independently or identically sampled from the same distribution, the local training loss (objective) functions can be disparate. This leads to different local objectives having different local minima, function values, and gradients. These new optimization challenges amplify the oscillating test performance phenomena discussed in Sec.\ref{sec:experiments}.

\subsection{Communication Model}
We assume bidirectional, synchronous, and noiseless device-to-device (D2D) communication channels that can be globally modeled as a flat, undirected, and connected communication graph with devices as vertices and D2D links as edges. Unlike federated \cite{mcmahan2017communication} or decentralized learning, there is no central server or coordinator. This implies that, apart from the complete graph, attaining global consensus \eqref{eq:consensus} requires multiple communication steps.
We further assume that data stays on-device and that privacy-preserving collaboration occurs only through device $k$ communicating its deep model parameters $w_k$ over D2D links.

\section{Algorithms}
\subsection{Peer-to-Peer Learning with Peer Affinity} \label{sec:our_alg}
Let $w_k^{(r,s,t)}$ be the $k$th device's model parameters in round $r$ after $s$ consensus steps and $t$ gradient updates. Let $w_k^{(r+1,0,0)}$, the parameters at the start of the next round, be equal to $w_k^{(r,S,T)}$, the parameters at the end of the previous round (with $S$ consensus and $T$ gradient steps). Let $w_k^{(0,0,0)}$ correspond to a random initialization (or the result of max norm synchronization in P2PL \cite{pranav2023peer}). Let $\alpha_{kj}$ be the mixing weight that device $k$ applies to neighbor $j$'s parameters such that $\forall k,j,\ \alpha_{kj} \geq 0$ and $\forall k,\ \alpha_{kk}+\sum_{j\in\mathcal{N}(k)}  \alpha_{kj} = 1$. Let $\mathcal{N}(k)$ be the $k$th device's set of neighbors and $\Tilde{\nabla}$ be a stochastic gradient. 
The P2PL with Affinity algorithm has two phases:\\
\noindent Peer-to-Peer Learning Phase:
\begin{equation}
    w_k^{(r,s,t+1)} = w_k^{(r,s,t)} - \eta \Tilde{\nabla}F_k\left(w_k^{(r,s,t)}\right) + \eta_d d^{(r,s,0)}_k\label{eq:our_grad_update}
\end{equation}
Consensus Phase:
\begin{align}
    w_k^{(r,s+1,t)} &= \alpha_{kk} w_k^{(r,s,t)} + \sum_{j\in\mathcal{N}(k)} \alpha_{kj} w_j^{(r,s,t)} + \eta_b b^{(r,0,t)}_k \label{eq:our_avg_consensus}
\end{align}
In \eqref{eq:our_grad_update} and \eqref{eq:our_avg_consensus}, $d^{(r,s,0)}_k$ and $b^{(r,0,t)}_k$ are bias terms that are designed to respectively drive learning and consensus phases to reduce oscillations in test performance. Hyperparameters $\eta_d$ and $\eta_b$ are the respective step sizes used for these bias terms. We note that $d^{(r,s,0)}_k$ is updated during a consensus step, but fixed during local learning; $b^{(r,0,t)}_k$ is updated during a local learning, but fixed during consensus steps. Like P2PL \cite{pranav2023peer}, \eqref{eq:our_grad_update} can include momentum, but we omit it here for simplicity.

One possible choice is $d_k^{(0,0,0)}=0$, $\eta_d = 1$, and the update $d^{(r,s+1,0)}_k = \frac{1}{T}\sum_{j\in\mathcal{N}(k)} \beta_{kj} \left(w_j^{(r,s,t)} - w_k^{(r,s,t)}\right)$, where $\forall k,j,\ \beta_{kj}>0$ and $\forall k,\ \sum_{j\in\mathcal{N}(k)}\beta_{kj}=1$. This corresponds to biasing device $k$ towards a weighted average of its neighbors during local training. Similarly, setting $b_k^{(0,0,0)}=w_k^{(0,0,0)}$, $\eta_b = 1$ and updating $b_k^{(r,0,t+1)} = \frac{1}{S}w_k^{(r,0,t+1)}$ corresponds to biasing device $k$ towards its parameters prior to consensus steps. When comparing to existing local-update algorithms, we note that updates to biases $d$ and $b$ do not require any additional communication.

\subsection{Existing Algorithms}
Distributed stochastic/mini-batch gradient descent (DSGD) is a popular peer-to-peer optimization algorithm that relies on alternating each local gradient descent update with a distributed average consensus step among neighboring devices. This corresponds to P2PL with Affinity with $S,T = 1$ and all biases $d,b = 0$. To reduce communication overhead, which is expensive in terms of runtime and power consumption, periodic/local-update DSGD (local DSGD) reduces communication frequency: each device executes multiple local gradient updates between average consensus steps. This corresponds to P2PL with Affinity with $S = 1$ and all biases $d,b = 0$.

Peer-to-Peer Learning (P2PL) \cite{pranav2023peer} is an algorithm that improves upon local DSGD by using an initial max-norm synchronization, momentum updates for local learning, row stochastic mixing weights, and device-specific consensus step sizes. 
The IID experiments in Sec.~\ref{sec:P2PL_IID_Data} use P2PL and the non-IID experiments in Sec.~\ref{sec:ours} use P2PL with Affinity. All other experiments in this paper use local DSGD to simplify analysis and discussion.


\section{Experiments}\label{sec:experiments}
All experiments involve devices training multilayer perceptrons (2NN \cite{mcmahan2017communication}) to classify MNIST \cite{lecun2010mnist} images of handwritten digits. We use PyTorch's \cite{Pytorch_NEURIPS2019_9015} default independent and random initialization followed by max-norm synchronization introduced in \cite{pranav2023peer}. Devices train using disjoint subsets of the 60,000 training samples and evaluate on 10,000 test samples.

\subsection{IID Data} \label{sec:P2PL_IID_Data}
In Ref.~\cite{pranav2023peer}, we introduced the Peer-to-Peer Learning (P2PL) algorithm and experimentally showed that, regardless of the connected communication graph, all devices in peer-to-peer settings achieve test performance attained by federated and centralized training. In this paper, we show that oscillations in test performance occur even when data at the devices is IID. 

Setup: Following \cite{pranav2023peer}, we train $K$ = 100 devices using P2PL. Local learning uses mini-batch ($B$ = 10) gradient descent with the PyTorch \cite{Pytorch_NEURIPS2019_9015} default variant of Polyak momentum, fixed $\eta$ = 0.01, and $\mu$ = 0.5. We randomly shuffle and equally partition the full training dataset into 100 local training datasets, where $\forall k,\ n_k = 600$. In each round, each device performs $60$ gradient updates before performing a single consensus step with neighboring devices. Like the experiments in Ref.~\cite{pranav2023peer}, we set $\forall t,k,\ \epsilon_k^{(t)}=1$ and set $\forall k,j,\ \alpha_{kj} = \frac{n_j}{n_k + \sum_{i\in\mathcal{N}(k)}n_i}$.  

\begin{figure}[t]
    \centering
    \includegraphics[width=0.66\columnwidth]{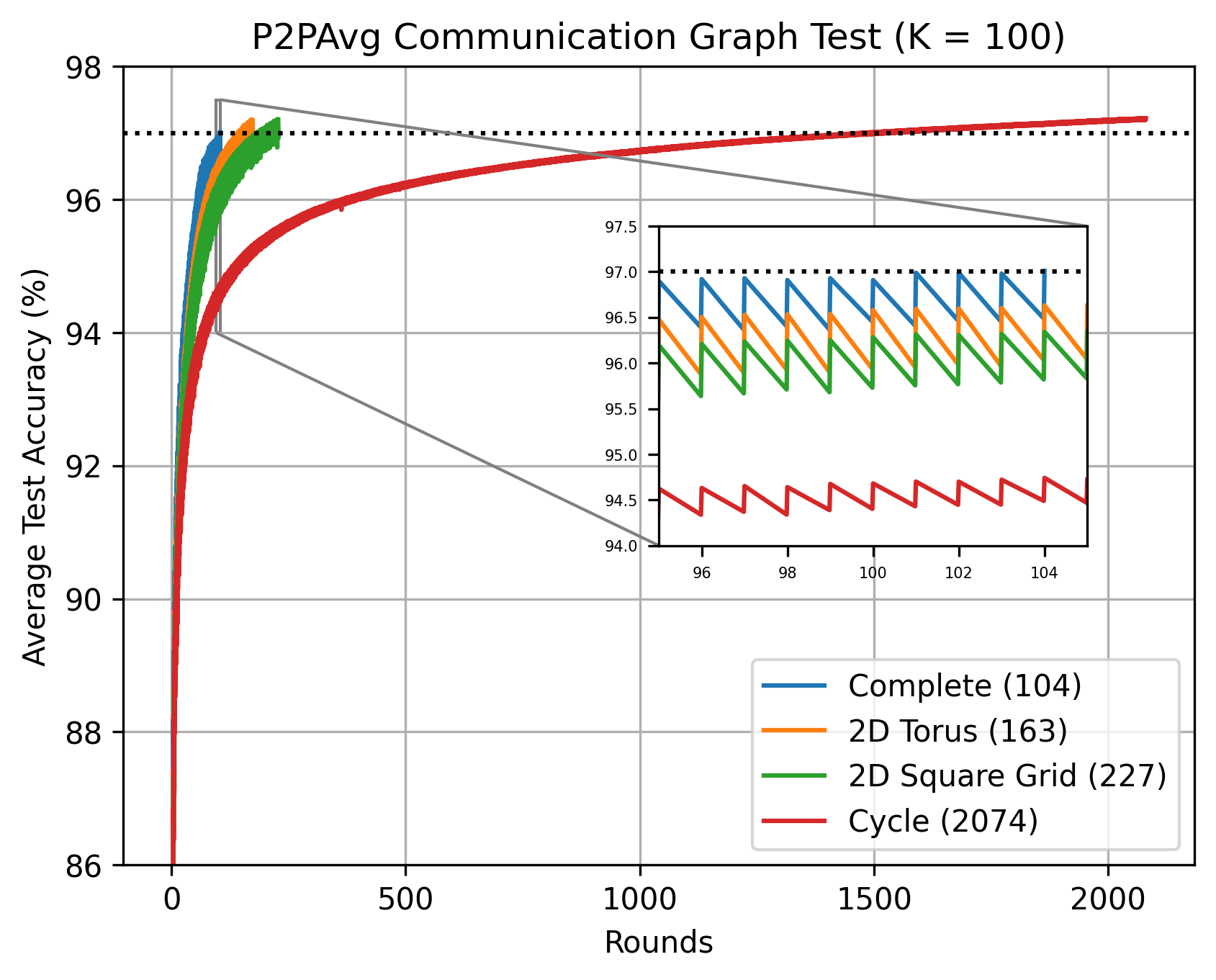}
    \caption{\fontsize{9}{10.5}\selectfont \textbf{P2PL convergence on various communication graphs with IID data} ($K$ = 100). Average test accuracy vs. rounds. Rounds to convergence in parentheses. Average accuracy may exceed $97\%$ before minimum accuracy. Drops in test accuracy correspond to local learning, while rises correspond to consensus steps.}
    \label{fig:IID_graph_test}
\end{figure}

Fig.~\ref{fig:IID_graph_test} extends the results in \cite{pranav2023peer} by evaluating test performance of all 100 deep models after local learning/training and again after the consensus phase of P2PL. This reveals that oscillations in test accuracy occur even in the IID setting. Although local learning decreases generalization performance (even when only a single local update is performed), test performance after the consensus phase surprisingly still shows a generally increasing trend across multiple rounds. Our observations reflect the intuition that local learning overfits to local data, but local learning is still necessary to extract gradient updates (knowledge) that can be shared in the consensus phase to improve neighbors' performance. The results highlight the importance of collaboration and consensus in every round and show that P2PL enables all devices to reach 97\% test accuracy despite these intermediate test performance oscillations.

\subsection{Non-IID Data}
Ref.~\cite{pranav2023peer} showed that, in non-IID data settings, Peer-to-Peer Learning (P2PL) enables devices to learn to classify data from classes that they never saw during training. This led to devices (in any connected communication graph) achieving test performance attained by federated and centralized training. In this paper, we attribute slower convergence in non-IID settings to larger oscillations in test performance evaluated after local learning and after consensus phases.

To isolate the effect of non-IID data on test performance oscillations, we simplify the problem to $K = 2$ devices.
This removes the effect of the communication graph topology on distributed consensus steps (global consensus is attained in 1 step) and allows us to clearly analyze the effect of ``pathological" non-IID data: where devices' local training datasets only have a subset of the classes seen at test time. Apart from the IID data partitioning, we follow the setup in Sec.~\ref{sec:P2PL_IID_Data}. We also set momentum $\mu$ = 0, effectively equating P2PL to local DSGD with max norm synchronization.

By removing confounding factors, 
we are able ask the more fundamental question: what happens when devices collaborate and learn to classify data from classes that they never saw during training (unseen classes)?

\subsubsection{Larger Oscillations and the Forgetting Problem} \label{sec:nonIID-oscillations}

\begin{figure}[t]
\centering
\begin{subfigure}[t]{0.49\linewidth}
  \centering
  
  \includegraphics[width=\textwidth]{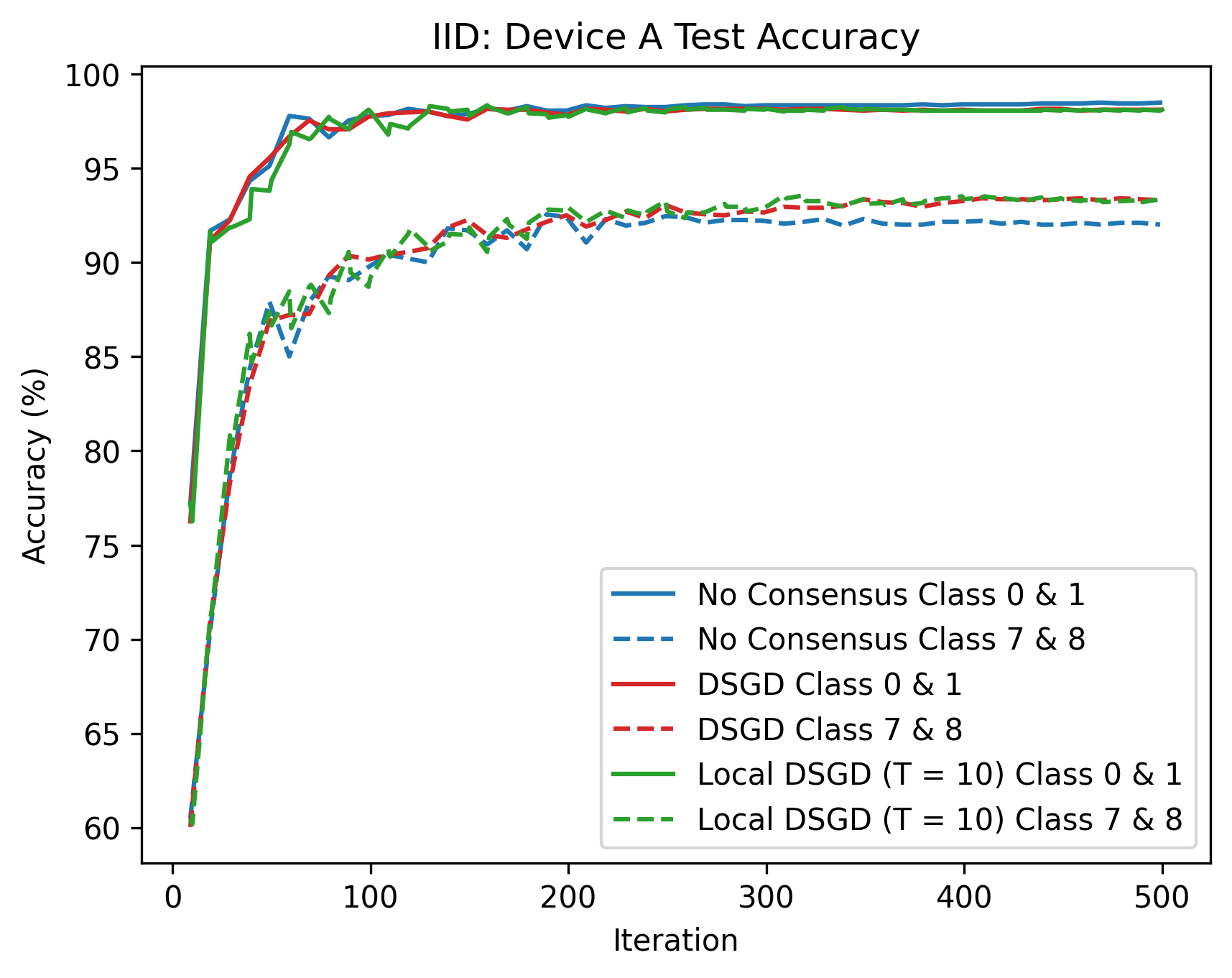}
  \caption{}
  \label{fig:IID-A}
\end{subfigure}\hfill
\begin{subfigure}[t]{0.49\linewidth}
  \centering
  
  \includegraphics[width=\textwidth]{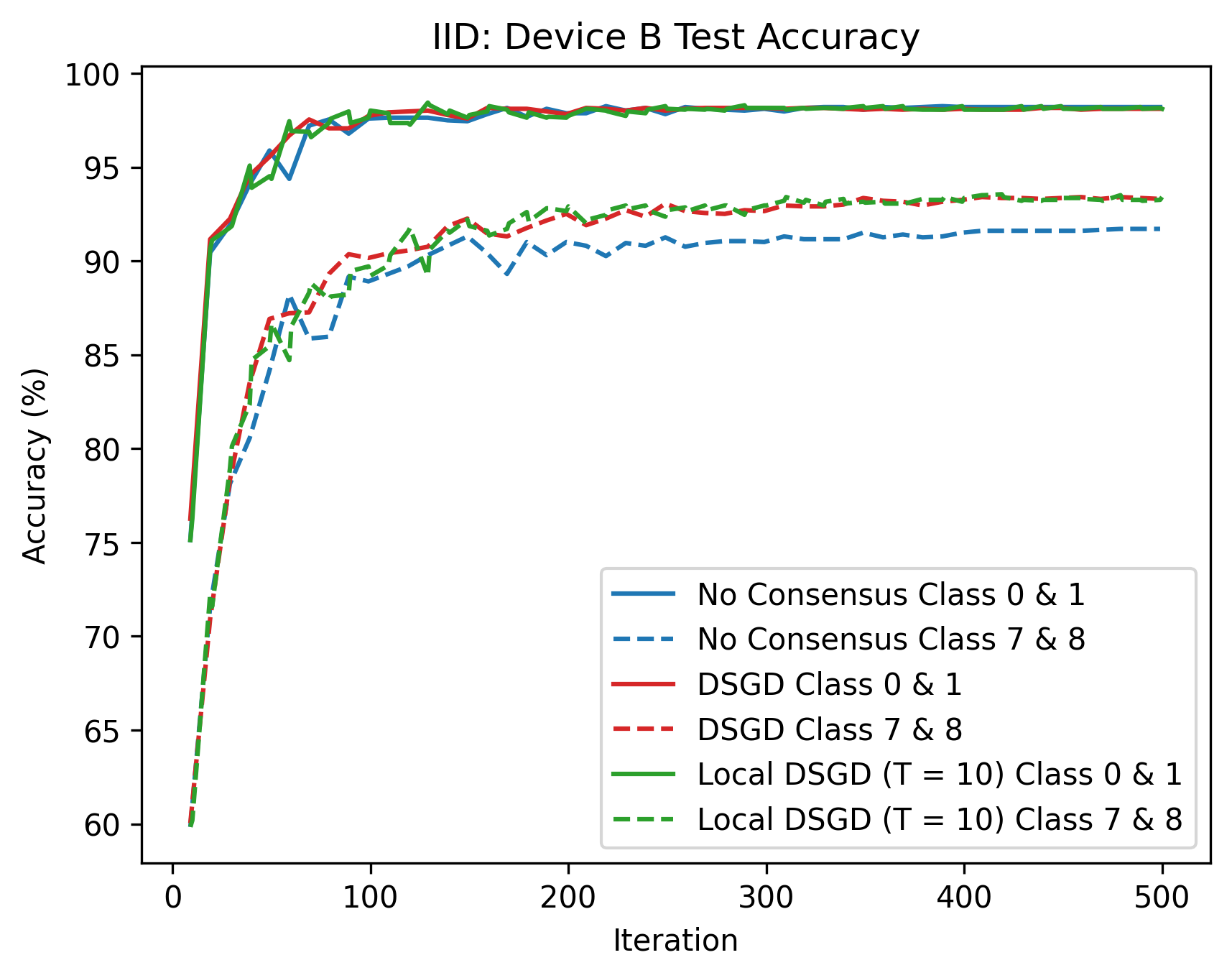}
  \caption{}
  \label{fig:IID-B}
\end{subfigure}\hfill
\begin{subfigure}[t]{0.49\linewidth}
  \centering
  
  \includegraphics[width=\textwidth]{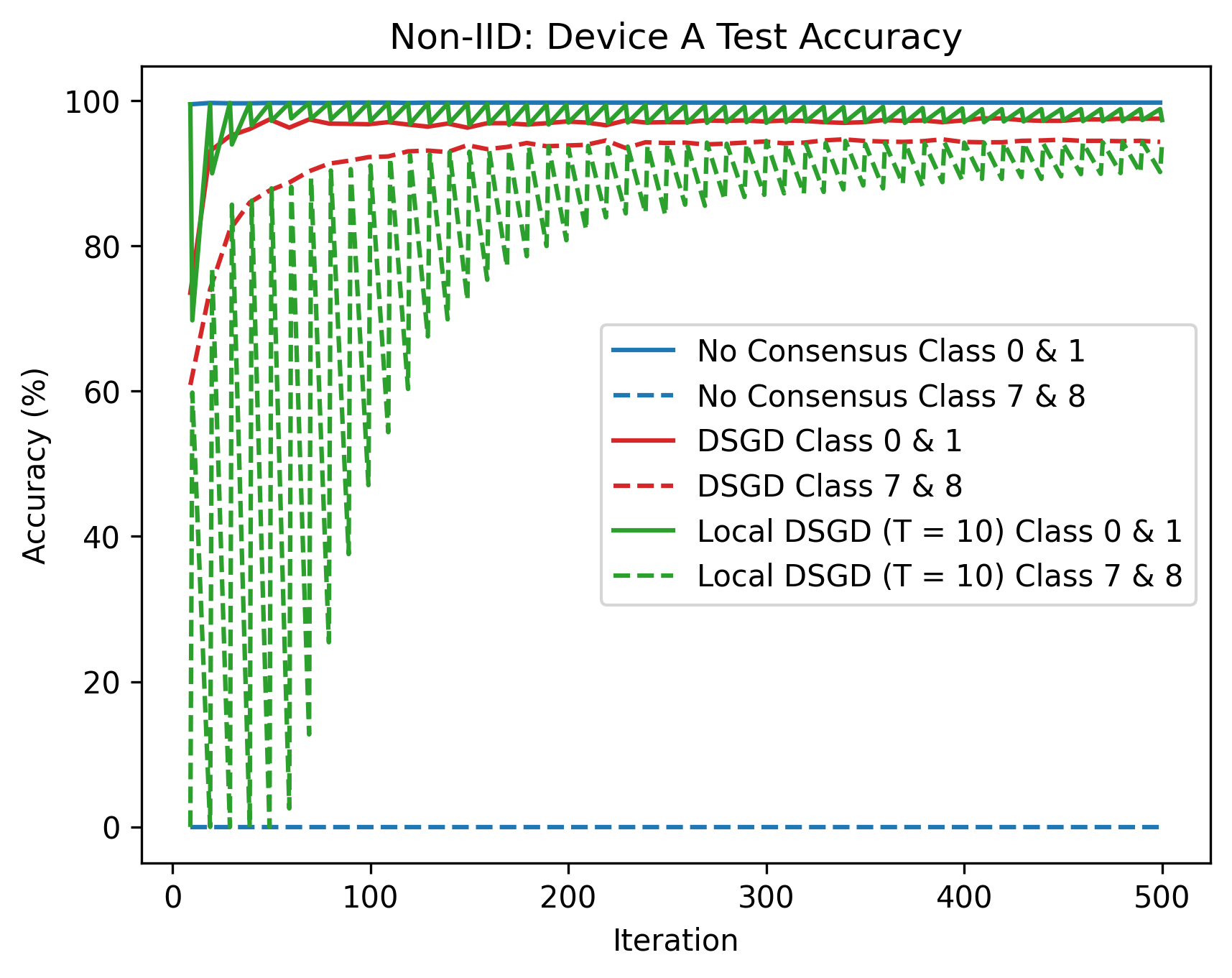} 
  \caption{}
  \label{fig:NonIID-A}
\end{subfigure}\hfill
\begin{subfigure}[t]{0.49\linewidth}
  \centering
  
  \includegraphics[width=\textwidth]{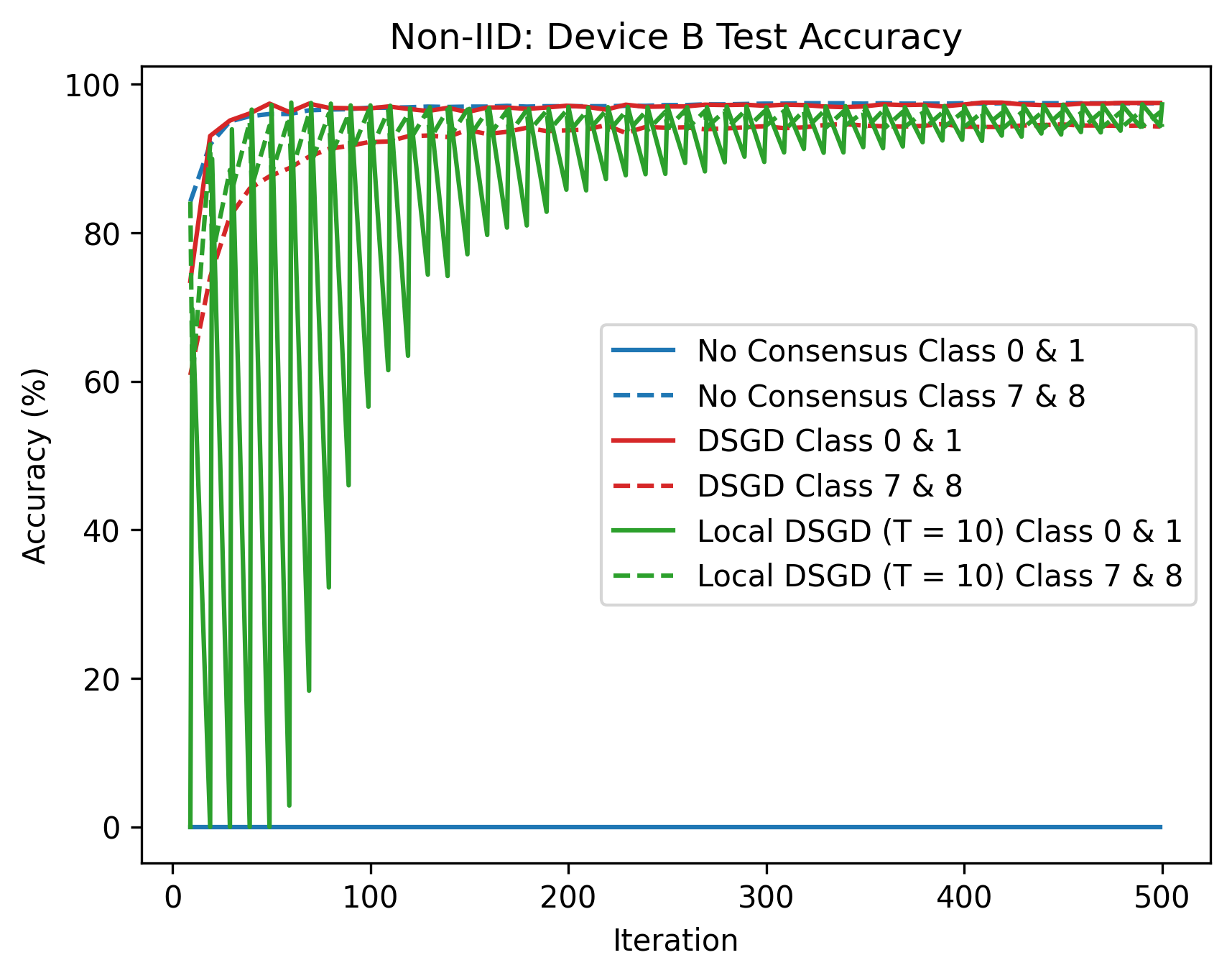}
  \caption{}
  \label{fig:NonIID-B}
\end{subfigure}
\caption{\fontsize{9}{10.5}\selectfont Stratified test accuracy vs. training iterations with IID and non-IID data. For IID experiments a) and b), both devices have access to 25 disjoint samples from each class 0, 1, 7, and 8. For c) and d), device A has 50 samples from each class 0 and 1 while device B has 50 samples from each class 7 and 8.}
\label{fig:small_nonIID}
\end{figure}

We examine the non-IID setting where device A trains on 50 samples from class 0 and 50 from class 1 while device B trains on 50 from class 7 and 50 from class 8. Both devices use batch size $B$ = 10. Fig.~\ref{fig:small_nonIID}cd show that local DSGD with consensus after 1 epoch results in both devices having drastic oscillations in test accuracy on unseen classes and smaller oscillations on seen classes.

Local training leads to a sharp drop in performance on unseen classes for local DSGD -- even to 0\% accuracy early in training. Unlike the IID setting in Fig.~\ref{fig:small_nonIID}ab, overfitting to local non-IID data leads to devices forgetting how to classify the unseen classes. Consensus steps then create a sharp increase in accuracy on unseen data to the point where local DSGD performance matches DSGD (which performs a consensus step after each gradient step). These results support the use of communication-efficient local DSGD-like algorithms and confirm that DSGD's performance on unseen classes acts as an upper bound on local DSGD's performance. As oscillations decay over time, local DSGD improves on unseen classes.

Meanwhile, local training creates smaller but significant increases in test performance on seen classes. The increases continue up to an upper bound corresponding to isolated local training without consensus. Each consensus step then drops performance on seen classes close to a lower bound corresponding to DSGD. These oscillations on seen classes also slowly decay as training progresses, but converge towards the lower bound of DSGD's performance.


\subsubsection{Number of Local Gradient Steps} \label{sec:num_gradient_steps}
\begin{figure}
\centering
\begin{subfigure}{0.49\linewidth}
  \centering
  
  \includegraphics[width=\textwidth]{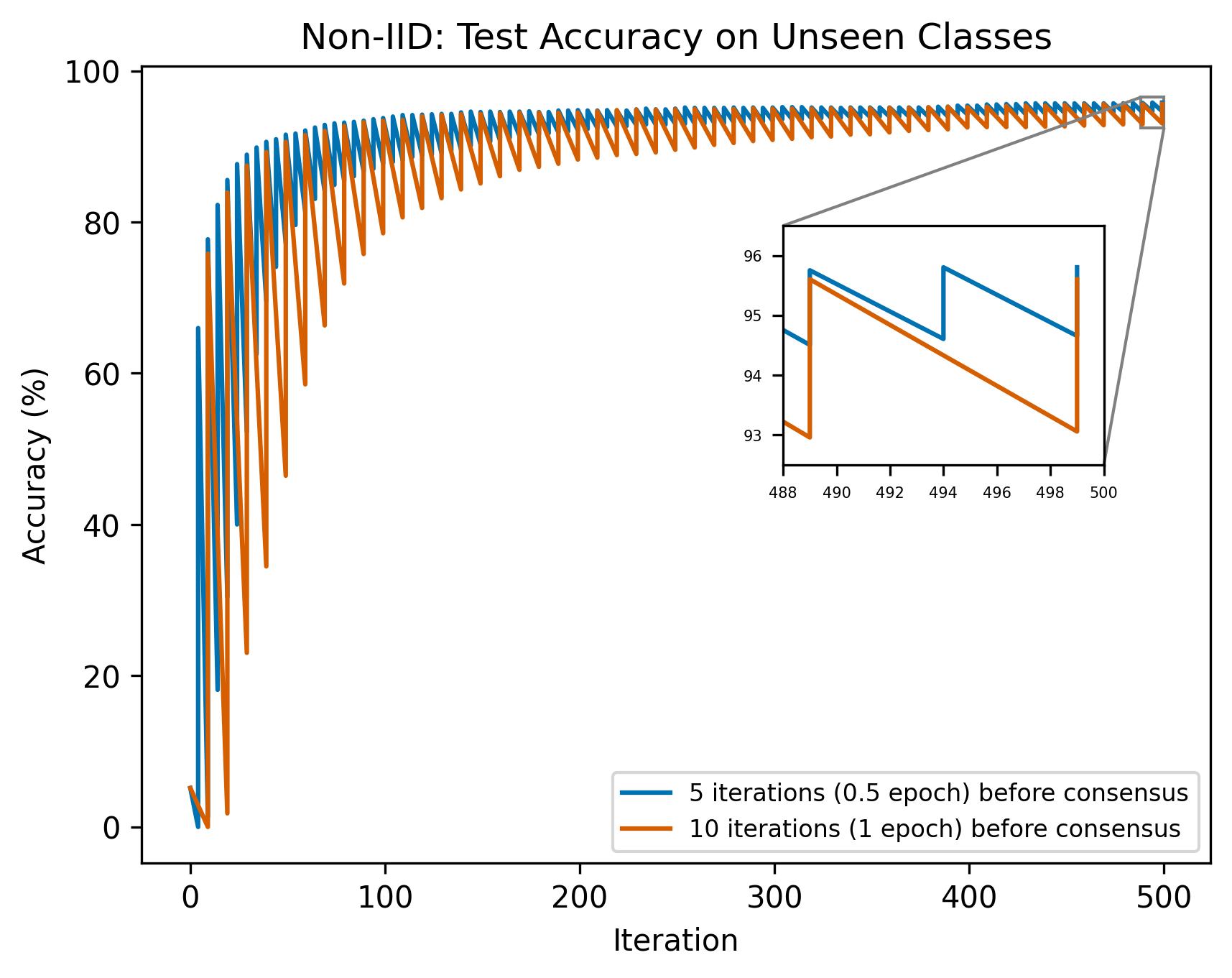}
  \caption{}
  \label{fig:unseen_local_iterations}
\end{subfigure}
\begin{subfigure}{0.49\linewidth}
  \centering
  
  \includegraphics[width=\textwidth]{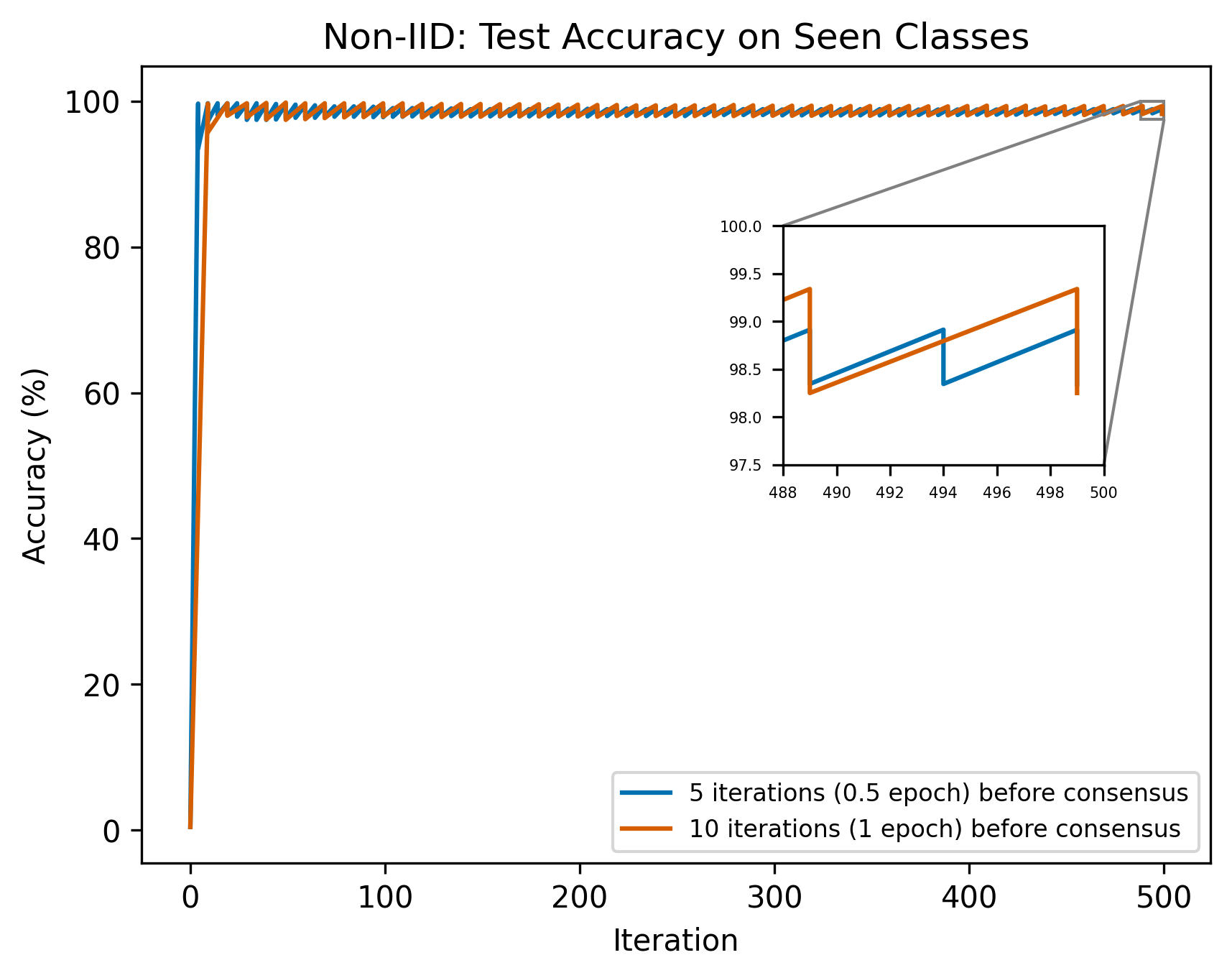}
  \caption{}
  \label{fig:seen_local_iterations}
\end{subfigure}
\caption{\fontsize{9}{10.5}\selectfont Plots are for device A trained with all samples from classes 0 and 1 using a batch size $B$ = 1240 (10 iter./epoch). a) Test accuracy on unseen classes 7 and 8. b) Test accuracy on seen classes 0 and 1.}
\label{fig:gradient_steps}
\end{figure}

Fig~\ref{fig:gradient_steps} shows that decreasing the number of local gradient steps, $T$, between consensus steps leads to smaller oscillations and noticeably higher test performance (0.2\%) on both seen and unseen data. However, the tradeoff for this small increase in accuracy is doubling the frequency and amount of communication. Since communication is expensive in practice due to power and latency issues, simply increasing communication does not provide efficient performance gains.

\subsubsection{Task Complexity}
\begin{figure}[t]
\centering
\begin{subfigure}[t]{0.49\linewidth}
  \centering
  
  \includegraphics[width=\textwidth]{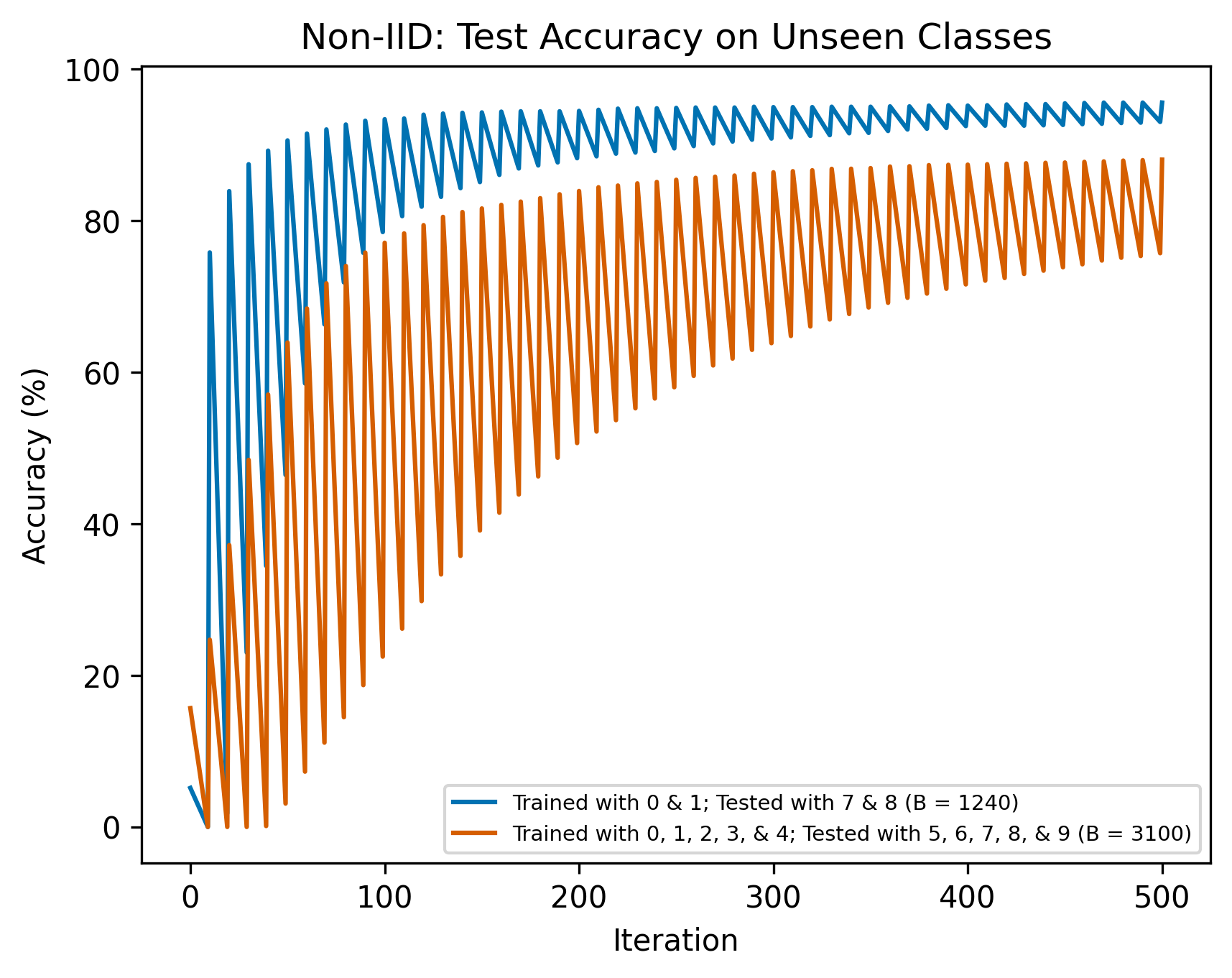}
  \caption{}
  \label{fig:unseen_num_classes}
\end{subfigure}
\begin{subfigure}[t]{0.49\linewidth}
  \centering
  
  \includegraphics[width=\textwidth]{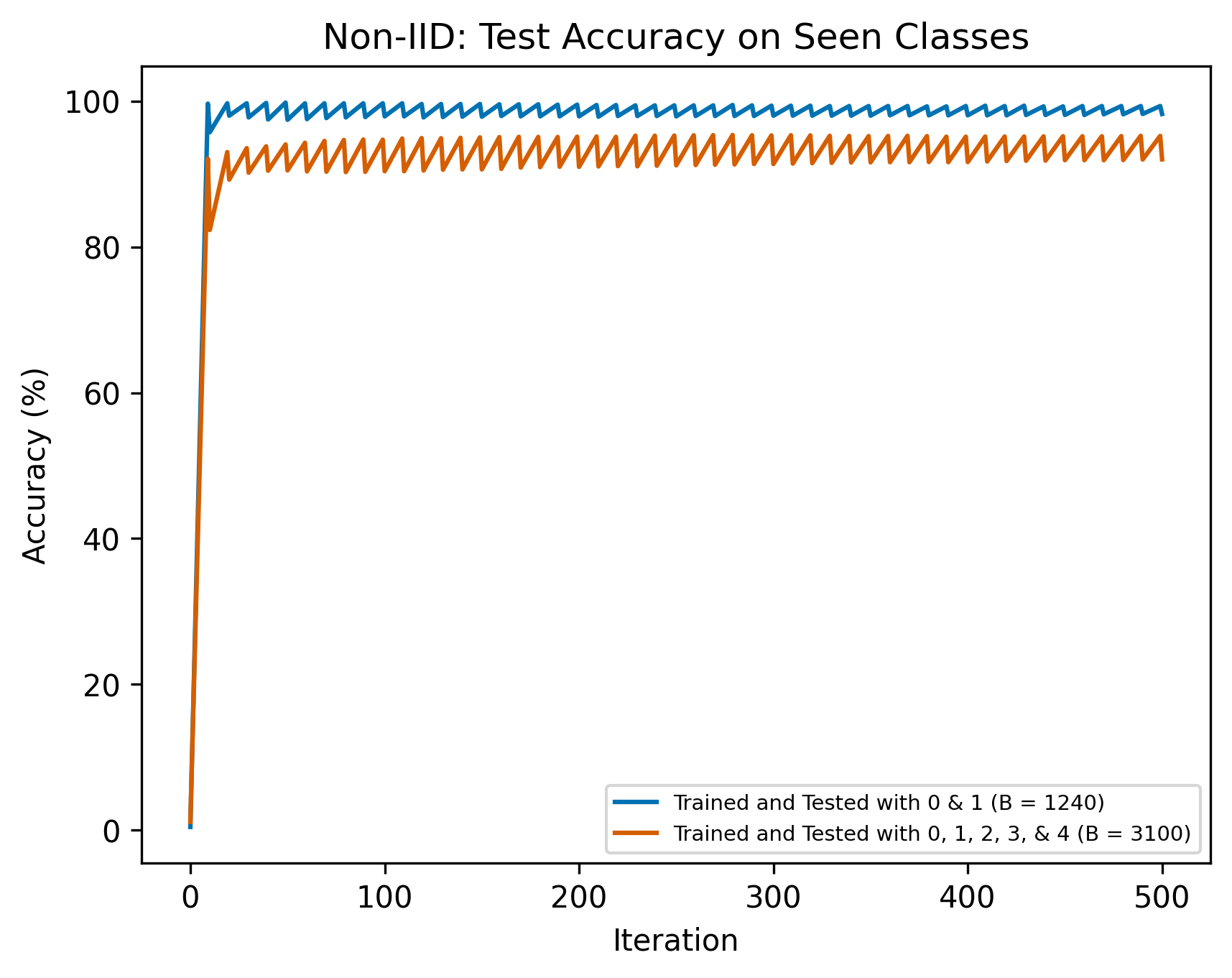}
  \caption{}
  \label{fig:seen_num_classes}
\end{subfigure}
\caption{ \fontsize{9}{10.5}\selectfont Plots are for device A trained with all samples from specified classes (with specified batch size such that there are 10 iter./epoch). a) Test accuracy on unseen classes. b) Test accuracy on seen classes.}
\label{fig:task_complexity}
\end{figure}

Fig.~\ref{fig:task_complexity} shows that increasing the complexity of the task from 4-class classification to 10-class classification leads to larger test performance oscillations. While the decrease in accuracy for both seen and unseen classes is expected with a more complex task, the larger oscillations indicate that the forgetting problem is more pronounced for harder tasks and damping oscillations in test performance is more important.
We note that the 10-class case has more local data, but uses a larger batch size such that 10 local iterations still corresponds to 1 epoch.

\subsection{Damping Oscillations using P2PL with Affinity}\label{sec:ours}
\begin{figure}[t]
\centering
\begin{subfigure}[t]{0.49\linewidth}
  \centering
  
  \includegraphics[width=\textwidth]{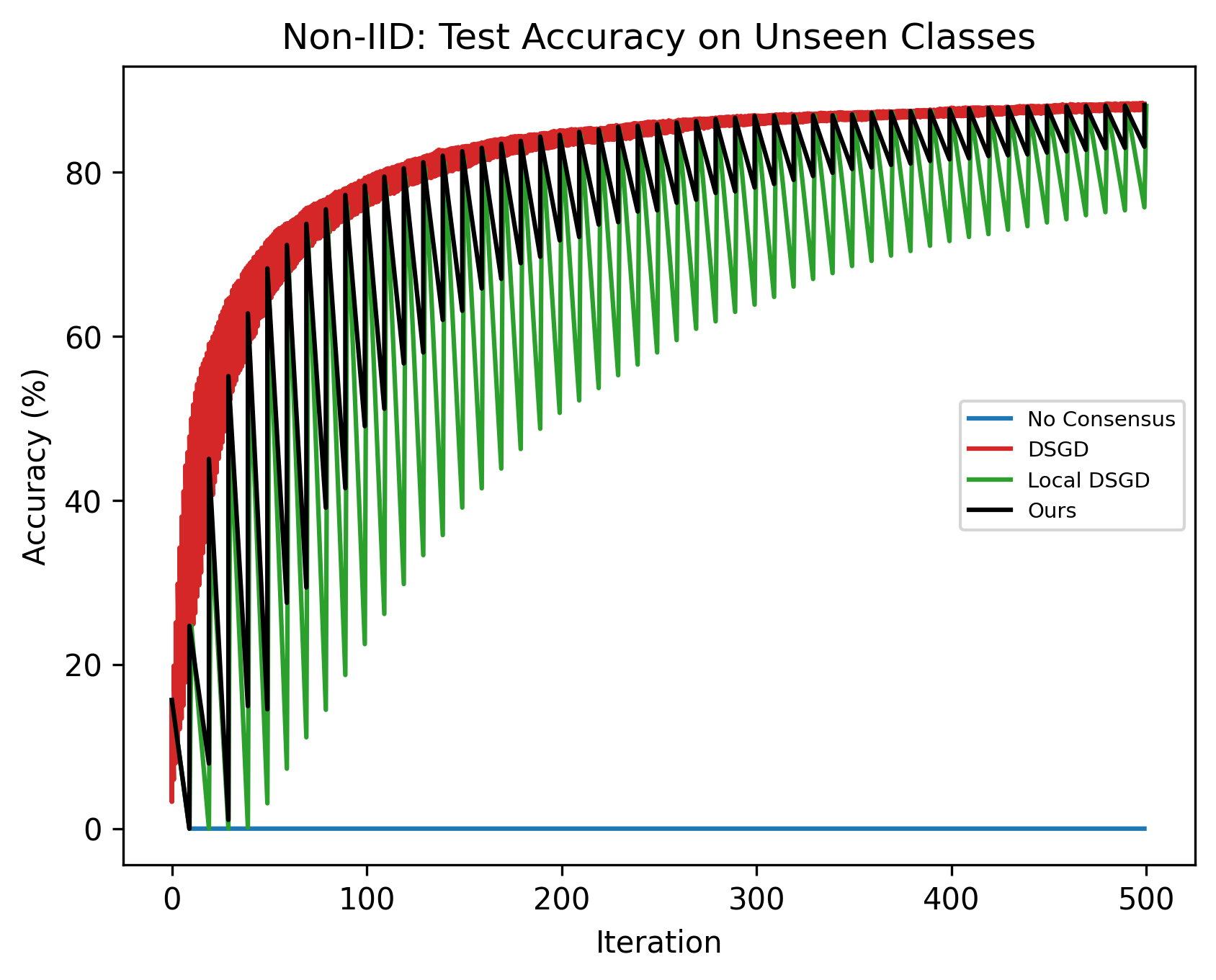}
  \caption{}
  \label{fig:unseen_algorithm_comparison}
\end{subfigure}
\begin{subfigure}[t]{0.49\linewidth}
  \centering
  
  \includegraphics[width=\textwidth]{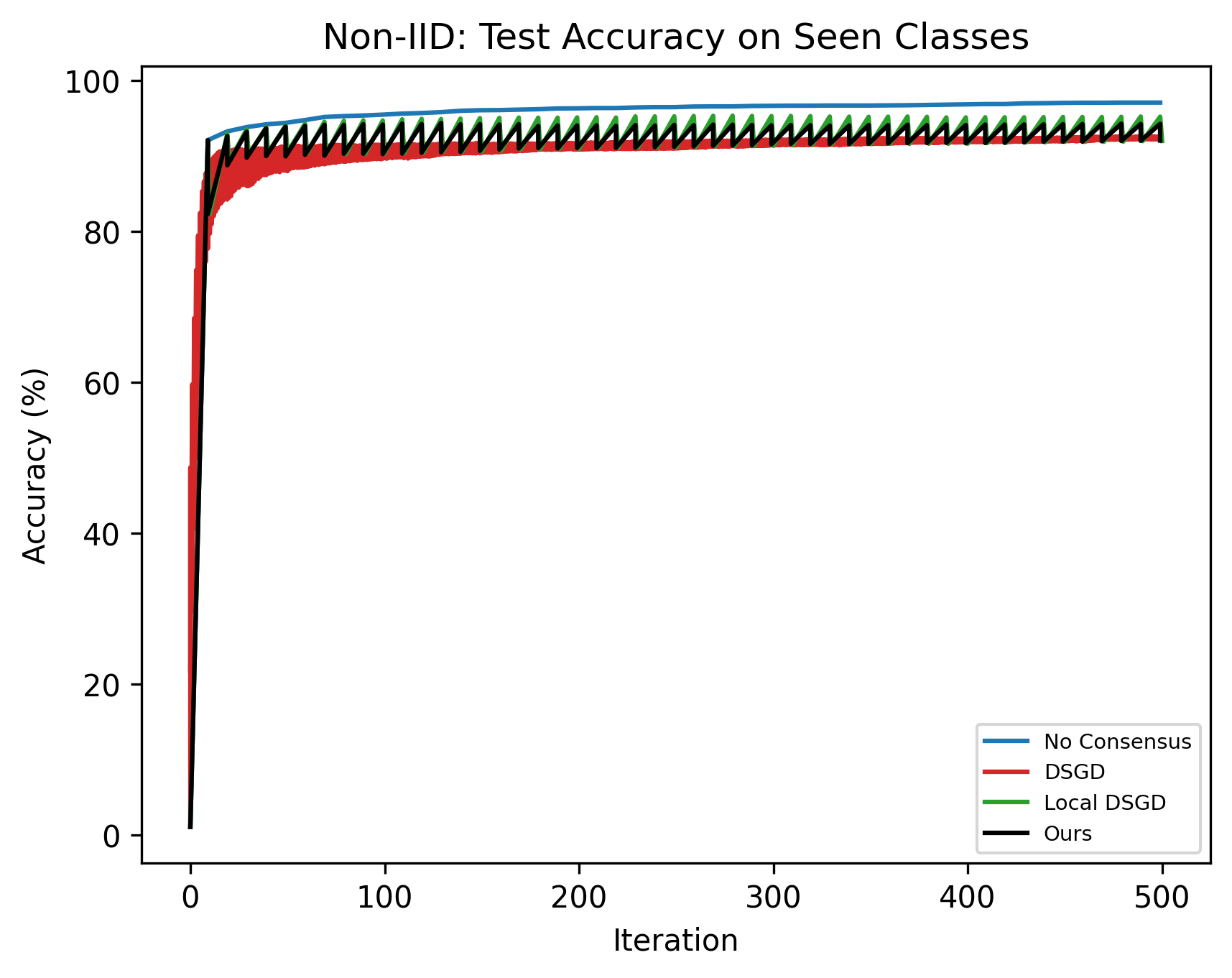}
  \caption{}
  \label{fig:seen_algorithm_comparison}
\end{subfigure}
\caption{\fontsize{9}{10.5}\selectfont Stratified test accuracy vs. training iterations for device A trained on classes 0, 1, 2, 3, and 4. a) Test accuracy on unseen classes 5, 6, 7, 8, and 9. b) Test accuracy on seen classes 0, 1, 2, 3, and 4.}
\label{fig:ours}
\end{figure}

We implement P2PL with Affinity as described in \ref{sec:our_alg}, except $\forall b,\ b = 0$ and $\beta_{kj} = \frac{n_j}{\sum_{i\in\mathcal{N}(k)} n_i}$. We compare with DSGD, local DSGD, and training without consensus. Fig.~\ref{fig:ours} shows that our proposed method successfully dampens oscillations on unseen data by reducing forgetting during local training and improving performance after consensus to match DSGD's performance. We attribute the improved performance to the local learning bias term $d$ that encourages device A's parameters to be close to its neighbors' parameters. With further fine-tuning of a nonzero bias $b$ in the consensus step, damping oscillations and improving performance on seen classes is also possible without additional communication overhead.

\section{Conclusion}
We show that drift/divergence in model parameters during local training creates oscillations in test performance evaluated after the local training phase and after the consensus phase.
These oscillations are present for P2PL and local DSGD-based algorithms in IID settings and they are further amplified by non-IID data, more local gradient steps, and higher task complexity. Experiments show that our proposed algorithm, P2PL with Affinity, dampens oscillations and improves performance by introducing bias terms in both algorithm phases.

  
  
  
  


\bibliographystyle{IEEEbib}
\tiny{
\bibliography{refs}
}

\end{document}